\documentclass[journal]{IEEEtran}
\usepackage{graphicx}
\usepackage{algorithm}
\usepackage{algorithmic}
\usepackage{booktabs,subcaption,amsfonts,dcolumn}
\usepackage{amstext}
\usepackage{amsmath}
\usepackage{amsfonts,amssymb}
\usepackage{amsmath,tabu}
\usepackage{amssymb}
\usepackage{array}
\usepackage{textcomp}
\usepackage{url}
\usepackage{verbatim}
\usepackage[dvips]{epsfig}
\usepackage{epsfig}
\usepackage[dvips]{graphicx}
\usepackage{multirow}
\usepackage{multirow} 
\usepackage[utf8]{inputenc}
\usepackage[english]{babel}
\usepackage{caption}
\usepackage{subcaption}
\usepackage{float}
\usepackage{ltablex}
\usepackage{cite}
\usepackage[table]{xcolor}
\usepackage{flushend}
\usepackage{setspace}

\usepackage{cleveref}
\usepackage[english]{babel}

\graphicspath{{D:/AsthaSaini/FCM_MM/IRW-FCM-TFS-main/code/results/}}

\begin{document}

\title{Comments on ``Iteratively Re-Weighted Algorithm for Fuzzy c-Means"}

\author{Astha Saini, Prabhu Babu
\thanks{Astha Saini and Prabhu Babu are with the Centre for Applied Research in Electronics, Indian Institute of Technology Delhi, New Delhi 110016, India (e-mail: astha.saini@care.iitd.ac.in, prabhu.babu@care.iitd.ac.in).}}



\maketitle

\begin{abstract}
   In this comment, we present a simple alternate derivation to the IRW-FCM algorithm presented in \cite{ref1} for Fuzzy c-Means problem. We show that the iterative steps derived in \cite{ref1} are nothing but steps of the popular Majorization Minimization (MM) algorithm. The derivation presented in this note is much simpler and straightforward and, unlike the derivation in \cite{ref1}, the derivation here does not involve introduction of any auxiliary variable. Moreover, by showing the steps of IRW-FCM as the MM algorithm, the inner loop of the IRW-FCM algorithm can be eliminated and the algorithm can be effectively run as a ``single loop" algorithm. More precisely, the new MM-based derivation deduces that a single inner loop of IRW-FCM is sufficient to decrease the Fuzzy c-means objective function, which speeds up the IRW-FCM algorithm.
\end{abstract}

\vspace{-1.5pt}
\section{Introduction and IRW-FCM Algorithm}
    Following the notations used in \cite{ref1}, let ${\mathbf{x}_{1},\mathbf{x}_{2},.....,\mathbf{x}_{n}}$ denote the data points that have to be divided into `$c$' clusters with centers $\{\mathbf{m}_j\}_{j=1}^c$, and $f_{ij}$ denote the $(i,j)^{\text{th}}$ element of the membership matrix $\mathbf{F}$ which indicates the membership of the $i^{\text{th}}$ data point to the $j^{\text{th}}$ cluster. It is obvious that the elements of $\mathbf{F}$ have to be non-negative and unlike the K-Means clustering technique, which requires the elements of $\mathbf{F}$ to be either 0 or 1, in the case of Fuzzy clustering the elements of $\mathbf{F}$ can be between 0 and 1 but the sum of elements in any row of $\mathbf{F}$ should be equal to 1, i.e., $\mathbf{F1}=\mathbf{1}$. With this, the Fuzzy c-Means problem can be stated as follows:
\begin{flalign}
\label{eqn1}
    &\min_{\mathbf {F1}=\mathbf{1},\mathbf F\geq 0,\{\mathbf{m}_j\}} \sum_{j=1}^{c} \sum_{i=1}^{n} f_{ij}^r \|\mathbf{x}_i - \mathbf{m}_j \|_2^2,& \nonumber\\
    &\min_{\mathbf {F1}=\mathbf{1},\mathbf F\geq 0,\{\mathbf{m}_j\}} \sum_{j=1}^{c} \sum_{i=1}^{n} f_{ij}^r (\mathbf{x}_i^T\mathbf{x}_i -2\mathbf{x}_i^T\mathbf{m}_j + \mathbf{m}_j^T\mathbf{m}_j).& 
\end{flalign}
    The parameter $r$ in \eqref{eqn1} indicates the level of fuzziness in the clustering. A usual approach to solve \eqref{eqn1} is to use alternating minimization, i.e., for a fixed $\mathbf{F}$, update \{$\mathbf{m}_j$\} and vice-versa. For a fixed $\mathbf{F}$, the minimizer over \{$\mathbf{m}_j$\} is given by:
\begin{equation}
\label{eqn2}
    \mathbf{m}_j = \frac{\sum_{i=1}^n f_{ij}^r \mathbf{x}_i}{\sum_{i=1}^n f_{ij}^r} = \frac{\sum_{i=1}^n g_{ij} \mathbf{x}_i}{\sum_{i=1}^n g_{ij}} =  \frac{\mathbf{X}\mathbf{g}_j}{\mathbf{g}_j^T\mathbf{1}},
\end{equation}
    where $g_{ij}=f_{ij}^r$. With fixed \{$\mathbf{m}_j$\}, solving for $\mathbf{F}$ via the Lagrange Multiplier method with the constraint $\sum_{j=1}^c f_{ij} =1$ gives: 
\begin{equation}
\label{eqn3}
    f_{ij} = \frac{d_{ij}^{(\frac{2}{1-r})}}{\sum_{k=1}^c d_{ik}^{(\frac{2}{1-r})}},
\end{equation}
    where $d_{ij}=\|\mathbf{x}_i - \mathbf{m}_j\|_2$. The alternating steps of \eqref{eqn2} and \eqref{eqn3} is known as FCM algorithm \cite{ref2}. In \cite{ref1}, the authors have stated that the FCM algorithm suffers from convergence to sub-optimal minimum by blaming the alternating approach of FCM, and they proposed a new algorithm named IRW-FCM, that minimizes an equivalent problem of \eqref{eqn1}. More precisely, they obtained a problem only in $\mathbf{F}$ by substituting for the optimal minimizer over \{$\mathbf{m}_j$\} (from \eqref{eqn2}) in \eqref{eqn1} and obtained the equivalent problem as shown below: 
\begin{flalign}
\label{eqn4}
    \min_{\mathbf{F1}=\mathbf{1},\mathbf F\geq 0} &\phi(\{\mathbf{g}_j\})  \stackrel{{\tiny \triangle}}{=} \nonumber \\
     &\sum_{j=1}^{c}\sum_{i=1}^{n} g_{ij} \mathbf{x}_i^T\mathbf{x}_i - \sum_{j=1}^{c} \left(\frac{\mathbf{g}_j^T \mathbf{X}^T \mathbf{X}\mathbf{g}_j}{\mathbf{g}_j^T\mathbf{1}}\right), 
\end{flalign}
    which they solved as follows. First, they introduced an auxiliary variable $\mathbf s \in \mathbb{R}^{c \times 1}$ and came up with a problem in $\{\mathbf{g}_j\}$ and $\mathbf{s}$, which is equivalent to \eqref{eqn4}:
\begin{equation}
\label{eqn5}
\begin{split}
    & \min_{\mathbf {F1}=\mathbf{1},\mathbf F\geq 0,\mathbf s} \psi(\{\mathbf{g}_j\},\mathbf{s}) \stackrel{{\tiny \triangle}}{=}   \\
    & \sum_{j=1}^{c}  \sum_{i=1}^{n} g_{ij} \mathbf{x}_i^T\mathbf{x}_i + \sum_{j=1}^{c} (s_j^2 \mathbf{g}_j^T\mathbf{1} - 2s_j\sqrt{\mathbf{g}_j^T\mathbf{X}^T\mathbf{X}\mathbf{g}_j}).
\end{split}
\end{equation}
    For a fixed $\mathbf{F}$, the optimal minimizer over $\mathbf{s}$ is given by $\left\{s_j=\frac{\sqrt{\mathbf{g}_j^T\mathbf{X}^T\mathbf{X}\mathbf{g}_j}}{\mathbf{g}_j^T\mathbf{1}}\right\}$, and substituting it back in \eqref{eqn5}, we get \eqref{eqn4} and this proves the equivalence of \eqref{eqn4} and \eqref{eqn5}. Then they proposed to solve \eqref{eqn5} via alternating minimization: For fixed $\mathbf{s}$, update $\mathbf{F}$ and vice-versa. For a fixed $\mathbf{s}$ (say $\mathbf{s}^{t+1}$ computed using $\mathbf{F}^t$), the minimization problem over $\mathbf{F}$ is given by:
\begin{equation}
\label{eqn6}
\begin{split}
    \min_{\mathbf {F1}=\mathbf{1},\mathbf F\geq 0} &\sum_{j=1}^{c}  \sum_{i=1}^{n} g_{ij} \mathbf{x}_i^T\mathbf{x}_i + \\
    &\sum_{j=1}^{c} \left((s_j^{t+1})^2 \mathbf{g}_j^T\mathbf{1} - 2s_j^{t+1}\sqrt{\mathbf{g}_j^T\mathbf{X}^T\mathbf{X}\mathbf{g}_j}\right).
\end{split}
\end{equation}
    However, \eqref{eqn6} does not have any closed form solution, so they resorted to another optimization strategy named IRW\cite{ref3}, which approximates the negative square root function $-\sqrt{\mathbf{g}_j^T\mathbf{X}^T\mathbf{X}\mathbf{g}_j}$ in \eqref{eqn6} with a linear function at some given $\mathbf{F}^t$, and solved the following approximated problem: 
\begin{equation} 
\label{eqn7}
    \min_{\mathbf {F1}=\mathbf{1},\mathbf F\geq 0} \sum_{j=1}^{c}  \sum_{i=1}^{n} g_{ij} \mathbf{x}_i^T\mathbf{x}_i + \sum_{j=1}^{c} ((s_j^{t+1})^2 \mathbf{1}^T\mathbf{g}_j - 2s_j^{t+1}\mathbf{a}_j^T\mathbf{g}_j),    
\end{equation}
    where $\mathbf{a}_j = \frac{\mathbf{X}^T\mathbf{X}\mathbf{g}_j^t}{\sqrt{(\mathbf{g}_j^t)^T\mathbf{X}^T\mathbf{X}\mathbf{g}_j^t}}$. The problem in \eqref{eqn7} has a closed form solution given by:
\begin{equation}
\label{eqn8}
    f_{ij}^{t+1} = \frac{(\mathbf{x}_i^T\mathbf{x}_i + (s_j^{t+1})^2 -2s_j^{t+1} a_{i j})^{\frac{1}{1-r}}}{\sum_{k=1}^{c}(\mathbf{x}_i^T\mathbf{x}_i + (s_k^{t+1})^2 -2s_k^{t+1} a_{i k})^{\frac{1}{1-r}}}.
\end{equation}
    Thus, the IRW-FCM approach which solves \eqref{eqn4} is iterative, and effectively it is a ``{\it{double loop}}" algorithm (one loop for alternating minimization and second inner loop for IRW procedure) as summarized in Algorithm \ref{alg:alg1}.
\begin{algorithm}[H]
\caption{IRW-FCM Algorithm}\label{alg:alg1}
    \begin{algorithmic}
        \STATE {\textbf{Input}}$(\mathbf{X}, c)$
        \STATE {\textbf{Initialize }} $\mathbf{F}^0$ 
        \STATE {\textbf{repeat}}
        \STATE \hspace{0.5cm} Calculate $\{s_j^{t+1}\}$ 
        \STATE \hspace{0.5cm} {\textbf{repeat}}
        \STATE \hspace{1 cm} Calculate $\mathbf{a}_j$
        \STATE \hspace{1 cm} Calculate $\mathbf{F}$ using equation \eqref{eqn8}
        \STATE \hspace{0.5cm} {\textbf{until} convergence}, {\textbf{Output}}  $\mathbf{F}^{t+1}$
        \STATE {\textbf{until} convergence}
        \STATE {\textbf{Output}} $\mathbf{F}^*$, $\mathbf{F}^*$ denotes the optimal minimizer.
    \end{algorithmic}
\label{alg1}
\end{algorithm}
\vspace{-0.8cm}
\section{MM BASED DERIVATION}
    The authors of \cite{ref1} criticised the alternating minimization approach of FCM by saying that it suffers from convergence to sub-optimal minimum and motivated to solve the equivalent problem in \eqref{eqn4} to avoid any such issues. However, in the process of solving \eqref{eqn4} they ``reversed back" and introduced a new auxiliary variable and again relied on the alternating minimization approach to solve the equivalent problem in \eqref{eqn5}. Moreover, they had to resort to a separate iterative procedure of IRW to solve one of the subproblems of the alternating minimization, which results in a ``double-loop" algorithm.\par
    
    In the following, we are going to present a much straightforward procedure in which we try and derive an iterative algorithm to solve \eqref{eqn4} by working only with the variable of interest ($\mathbf{F}$). We will employ a popular approach named Majorization-Minimization (MM)\cite{ref4} to accomplish this. Before we venture into solving the problem in \eqref{eqn4}, we first briefly introduce the MM approach. Let us say that we are interested in minimizing a function $f(\mathbf{x})$ over a domain $\mathbf{\chi}$, MM does the job in two steps: In the first step, given some $\mathbf{x}^t$ ($\mathbf{x}^t \in \mathbf{\chi}$), it constructs a tighter upperbound $h(\mathbf{x}|\mathbf{x}^t)$ to the cost function $f(\mathbf{x})$, and in the second step it minimizes the upperbound $h(\mathbf{x}|\mathbf{x}^t)$ w.r.t. $\mathbf{x}$ and obtains the next iterate $\mathbf{x}^{t+1}$ and this is repeated till convergence. The upperbound $h(\mathbf{x}|\mathbf{x}^t)$ should satisfy the following properties:
\begin{equation}
\label{eqn9}
    f(\mathbf{x}^t) = h(\mathbf{x}^t|\mathbf{x}^t),\text{and} \: f(\mathbf{x}) \leq h(\mathbf{x}|\mathbf{x}^t)\;\; \forall \,\mathbf{x} \in \mathbf{\chi} .
\end{equation}    
    It is easy to verify that the series of iterates obtained via MM will monotonically decrease the objective as shown below:
\begin{equation}
\label{eqn10}
    f(\mathbf{x}^{t+1})\leq h(\mathbf{x}^{t+1}|\mathbf{x}^t)\leq h(\mathbf{x}^t|\mathbf{x}^t)=f(\mathbf{x}^t).
\end{equation}
    The first inequality in \eqref{eqn10} is due to \eqref{eqn9} and the second inequality is due to fact that $\mathbf{x}^{t+1}$ is obtained by minimizing $h(\mathbf{x}|\mathbf{x}^t)$, and the last equality is also due to \eqref{eqn9}. Now, coming back to problem in \eqref{eqn4}, it is easy to see that the second term (without the negative sign) in \eqref{eqn4} is a quadratic over linear function (with $\mathbf{g}_j^T\mathbf{1} > 0$), which is a convex function (see section 3.1.5 in \cite{ref5}) or, in other words, the function $-\frac{\mathbf{g}_j^T \mathbf{X}^T \mathbf{X}\mathbf{g}_j}{\mathbf{g}_j^T\mathbf{1}}$ is a concave function in `$\mathbf{g}_j$'.  Thus a tighter upperbound (at a given $\mathbf{F}^t$) can be obtained via a tangent hyperplane passing through $\mathbf{F}^t$, which of course can be obtained via first order Taylor series expansion as shown below:
\begin{equation}
\label{eqn11}
\begin{split}
    & -\frac{\mathbf{g}_j^T \mathbf{X}^T \mathbf{X}\mathbf{g}_j}{\mathbf{g}_j^T\mathbf{1}} \leq -\frac{(\mathbf{g}_j^t)^T \mathbf{X}^T\mathbf{X}\mathbf{g}_j^t}{(\mathbf{g}_j^t)^T\mathbf{1}} \;- \\ 
    &\bigg{[}\frac{2(\mathbf{g}_j^t)^T\mathbf{1}(\mathbf{g}_j^t)^T\mathbf{X}^T\mathbf{X} - (\mathbf{g}_j^t)^T\mathbf{X}^T\mathbf{X}\mathbf{g}_j^t\mathbf{1}^T}{\mathbf{1}^T\mathbf{g}_j^t(\mathbf{g}_j^t)^T\mathbf{1}}(\mathbf{g}_j - \mathbf{g}_j^t)\bigg{]},
\end{split}    
\end{equation}   
    with equality achieved at $\mathbf{g}_j = \mathbf{g}_j^t$.The MM upperbound for the objective in \eqref{eqn4} at some given $(\mathbf{F}^t)$ is then given by:

\begin{flalign}
\label{eqn12}
    &\phi(\{\mathbf{g}_j\}) \leq \sum_{j=1}^{c}\sum_{i=1}^{n} g_{ij} \mathbf{x}_i^T\mathbf{x}_i - \sum_{j=1}^{c}\bigg{[}\frac{(\mathbf{g}_j^t)^T \mathbf{X}^T\mathbf{X}\mathbf{g}_j^t}{(\mathbf{g}_j^t)^T\mathbf{1}} \; +& \nonumber \\
    &\frac{2(\mathbf{g}_j^t)^T\mathbf{1}(\mathbf{g}_j^t)^T\mathbf{X}^T\mathbf{X} - (\mathbf{g}_j^t)^T\mathbf{X}^T\mathbf{X}\mathbf{g}_j^t\mathbf{1}^T}{\mathbf{1}^T\mathbf{g}_j^t(\mathbf{g}_j^t)^T\mathbf{1}}(\mathbf{g}_j - \mathbf{g}_j^t)\bigg{]}& \nonumber \\    
    &\stackrel{{\tiny \triangle}}{=} h(\{\mathbf{g}_j\}|\{\mathbf{g}_j^t\}).& 
\end{flalign}   
 
Then, the upperbound minimization problem to be solved is given by:
\begin{equation}
\label{eqn13}
\begin{split}
    &\min_{\mathbf {F1}=\mathbf{1},\mathbf F\geq 0} h(\{\mathbf{g}_j\}|\{\mathbf{g}_j^t\}) = 
    \sum_{j=1}^{c}\sum_{i=1}^{n} g_{ij} \mathbf{x}_i^T\mathbf{x}_i - \\
    &\sum_{j=1}^{c}\bigg{[} 2\frac{(\mathbf{g}_j^t)^T \mathbf{X}^T \mathbf{X}\mathbf{g}_j}{(\mathbf{g}_j^t)^T\mathbf{1}} -\frac{(\mathbf{g}_j^t)^T \mathbf{X}^T \mathbf{X}\mathbf{g}_j^t\mathbf{1}^T\mathbf{g}_j}{\mathbf{1}^T\mathbf{g}_j^t(\mathbf{g}_j^t)^T\mathbf{1}}\bigg{]} + \text{const.}
\end{split}
\end{equation}
Leaving out the constant terms, we get 
\begin{equation}
\label{eqn14}
\begin{split}
  \min_{\mathbf {F1}=\mathbf{1},\mathbf F\geq 0}  &\sum_{j=1}^{c}\sum_{i=1}^{n} g_{ij} \mathbf{x}_i^T\mathbf{x}_i\; -\\
  & \sum_{j=1}^{c} \bigg{[} 2\frac{(\mathbf{g}_j^t)^T \mathbf{X}^T \mathbf{X}\mathbf{g}_j}{(\mathbf{g}_j^t)^T\mathbf{1}} -\frac{(\mathbf{g}_j^t)^T \mathbf{X}^T \mathbf{X}\mathbf{g}_j^t\mathbf{1}^T\mathbf{g}_j}{\mathbf{1}^T\mathbf{g}_j^t(\mathbf{g}_j^t)^T\mathbf{1}}\bigg{]},
\end{split}
\end{equation}
    which has a simple closed form solution given by:
\begin{equation}
\label{eqn15}
    f_{ij}^{t+1} =  \frac{\left[\mathbf{x}_i^T\mathbf{x}_i + \frac{(\mathbf{g}_j^t)^T \mathbf{X}^T \mathbf{X}\mathbf{g}_j^t}{\mathbf{1}^T\mathbf{g}_j^t(\mathbf{g}_j^t)^T\mathbf{1}} -2\frac{\mathbf{x}_i^T\mathbf{X}\mathbf{g}_j^t}{(\mathbf{g}_j^t)^T\mathbf{1}}\right]^{\frac{1}{1-r}}}{\sum_{k=1}^{c}\left[\mathbf{x}_i^T\mathbf{x}_i + \frac{(\mathbf{g}_k^t)^T \mathbf{X}^T \mathbf{X}\mathbf{g}_k^t}{\mathbf{1}^T\mathbf{g}_k^t(\mathbf{g}_k^t)^T\mathbf{1}} -2\frac{\mathbf{x}_i^T\mathbf{X}\mathbf{g}_k^t}{(\mathbf{g}_k^t)^T\mathbf{1}}\right]^{\frac{1}{1-r}}} \forall{i,j}.
\end{equation}   
    If one takes a closer look at the problem in \eqref{eqn14}, it is nothing but the problem in \eqref{eqn7} with choice of $s_j^{t+1} =\frac{\sqrt{(\mathbf{g}_j^t)^T\mathbf{X}^T\mathbf{X}\mathbf{g}_j^t}}{(\mathbf{g}_j^t)^T\mathbf{1}}$, $ \mathbf{a}_j = \frac{\mathbf{X}^T\mathbf{X}\mathbf{g}_j^t}{\sqrt{(\mathbf{g}_j^t)^T\mathbf{X}^T\mathbf{X}\mathbf{g}_j^t}}$. This in a way proves that the iterative steps of IRW-FCM can be easily derived via the MM approach, and more importantly, the inner loop of IRW-FCM algorithm can be run only for one time and that is sufficient to decrease the cost function in \eqref{eqn4}.

\begin{table}[!ht]
\captionsetup{font=small}
\centering
\begin{tabular}{|p{1.1cm}|p{3.35cm}|p{3.35cm}|}
 \hline
 \textbf{Sr. No.}& \textbf{FCM-MM} & \textbf{FCM-IRW}\\
 \hline
1. & Single loop algorithm. & Double loop algorithm. \\
 \hline
2. & Does not involve  introduction of any auxiliary variable therefore has less space complexity. & Requires auxiliary variables and therefore requires more storage.  \\
 \hline
3. & Faster than FCM-IRW as it is equivalent to FCM-IRW with the inner loop run only one time. & Slower due to the redundant inner loop. \\
 \hline
 \end{tabular}
\caption{DIFFERENCES AND SIMILARITIES BETWEEN FCM-MM AND FCM-IRW ALGORITHMS}
\label{table:tab1}
\end{table}
\section{Numerical Simulation Results}
\begin{figure*}[!ht]
\captionsetup{font=small}
\centering
    \begin{subfigure}{0.19\textwidth}
        \centering
        \includegraphics[width=\textwidth,height=1.5cm]{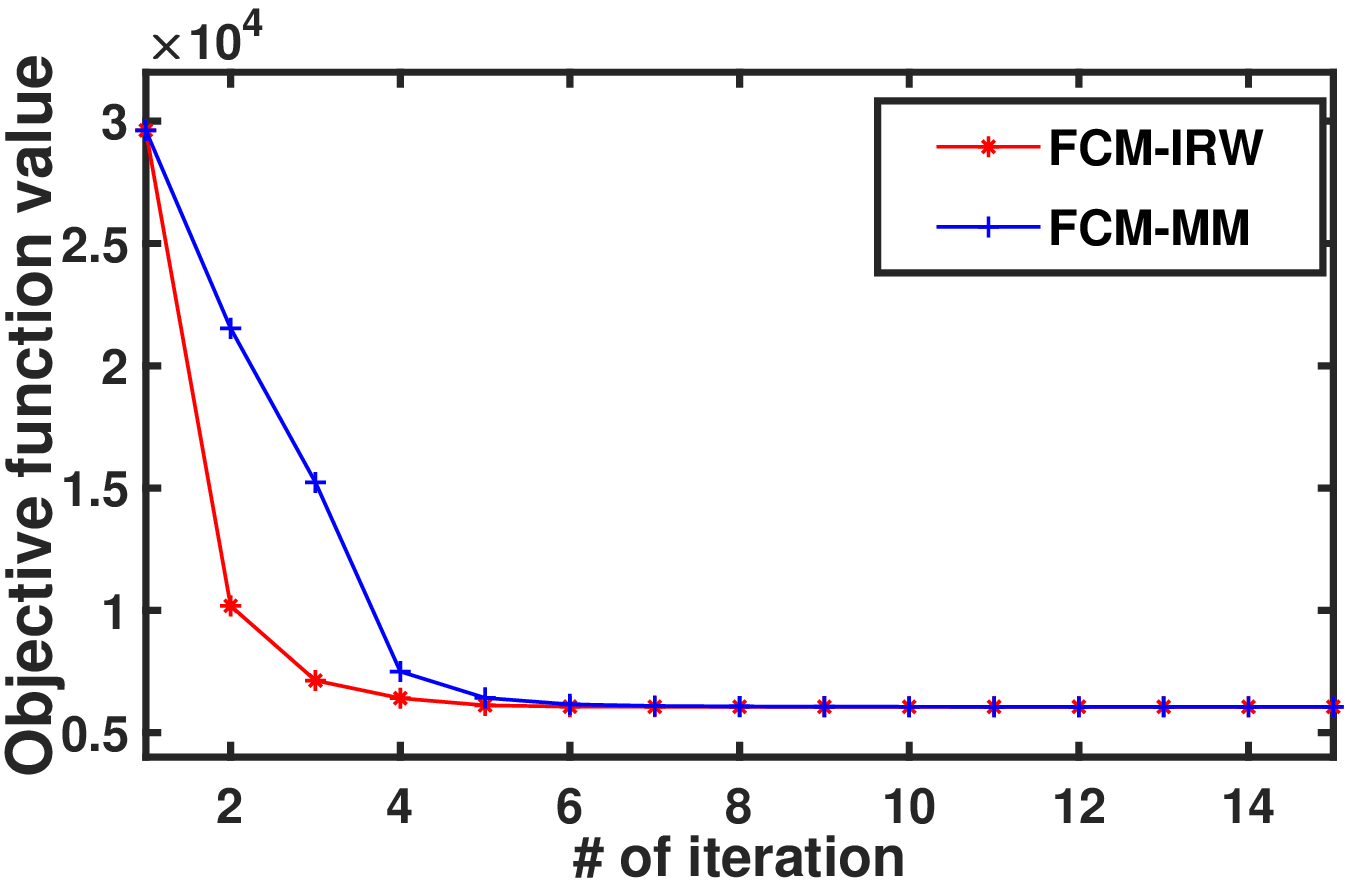}
        \caption{}
        \label{fig:1.a}
    \end{subfigure}
    \hfill
    \begin{subfigure}{0.19\textwidth}
        \centering
        \includegraphics[width=\textwidth,height=1.5cm]{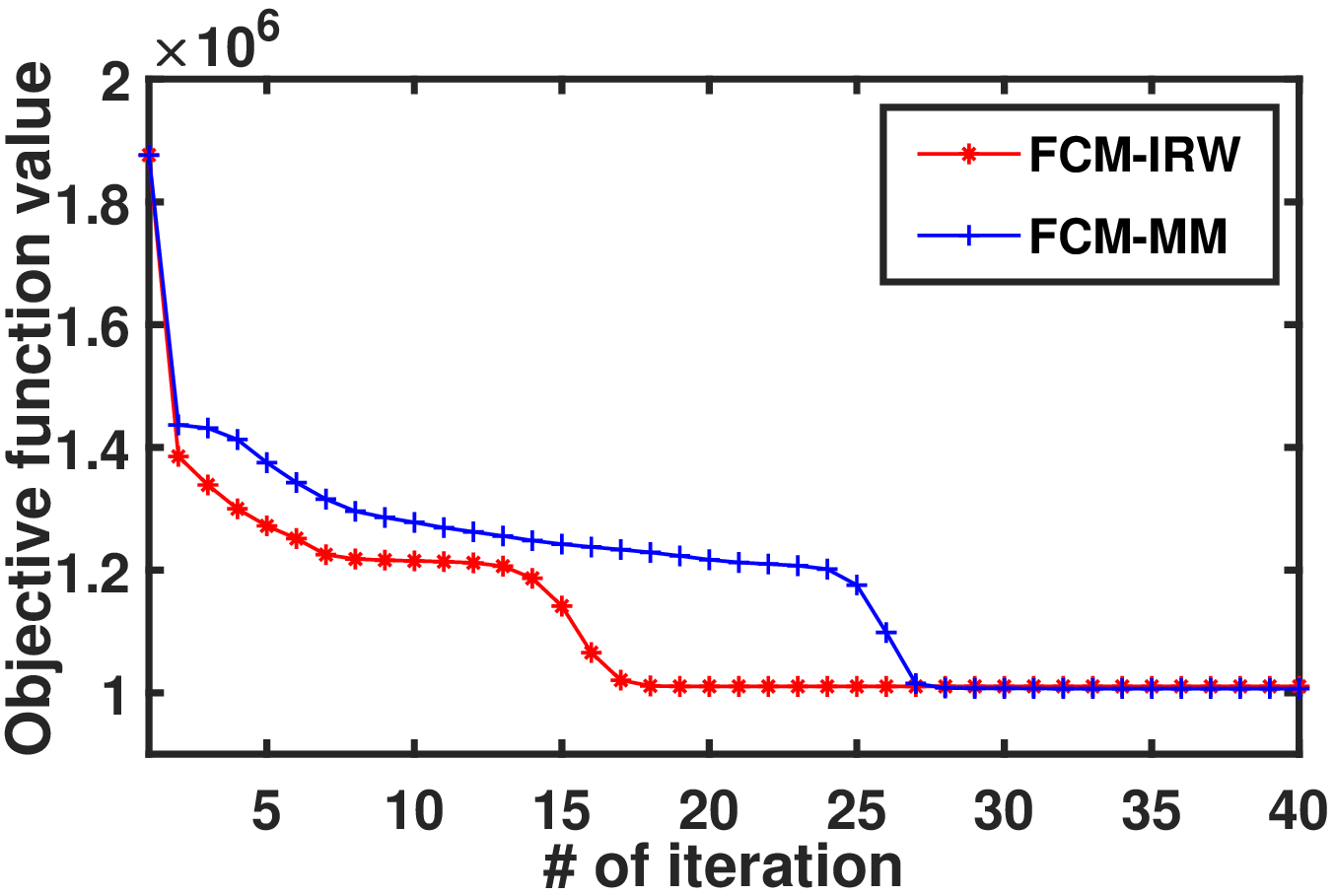}
        \caption{}
        \label{fig:1.b}
    \end{subfigure}
    \hfill
    \begin{subfigure}{0.19\textwidth}
        \centering
        \includegraphics[width=\textwidth,height=1.5cm]{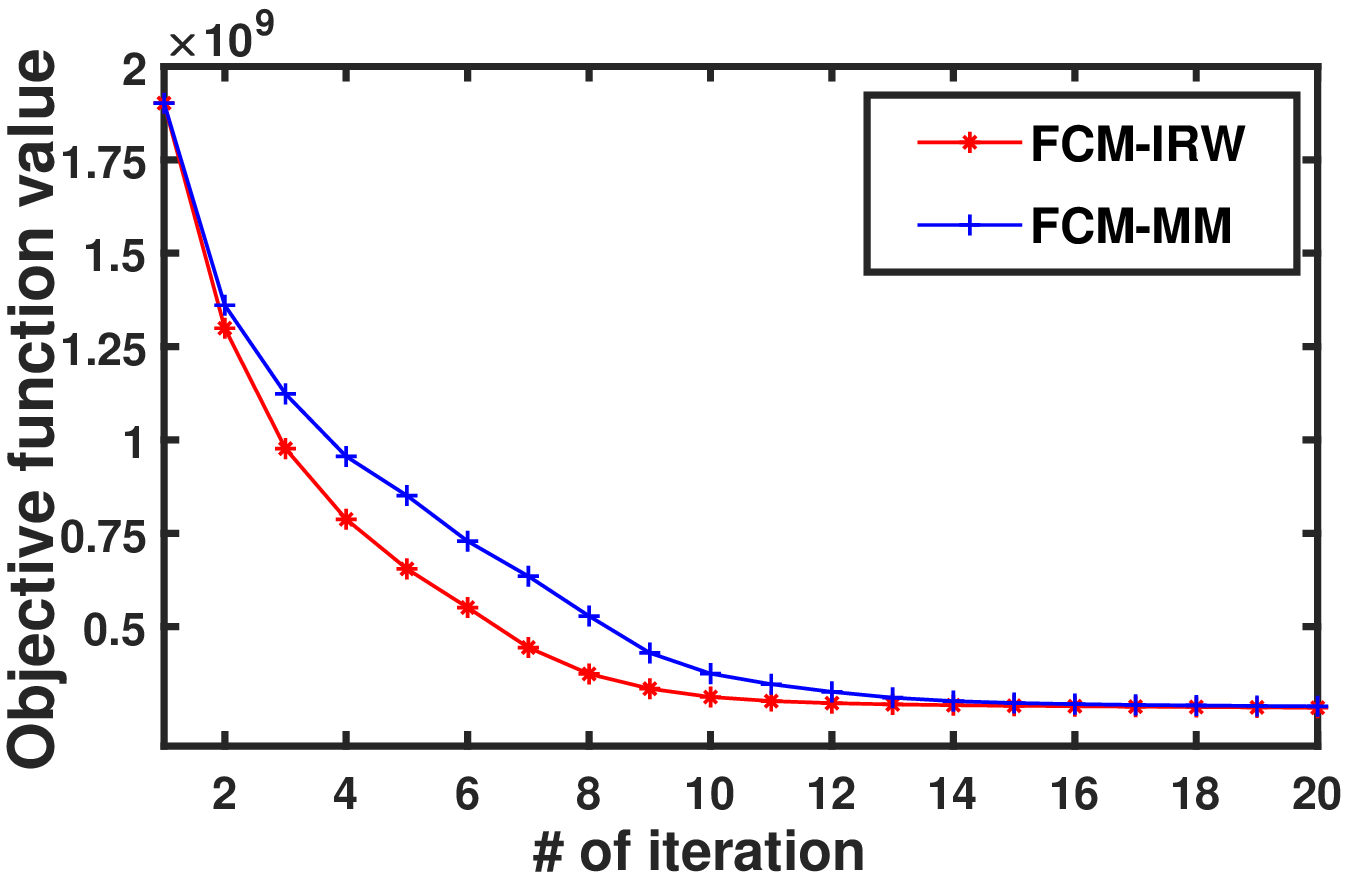}
        \caption{}
        \label{fig:1.c}
    \end{subfigure}
    \hfill
    \begin{subfigure}{0.19\textwidth}
        \centering
        \includegraphics[width=\textwidth,height=1.5cm]{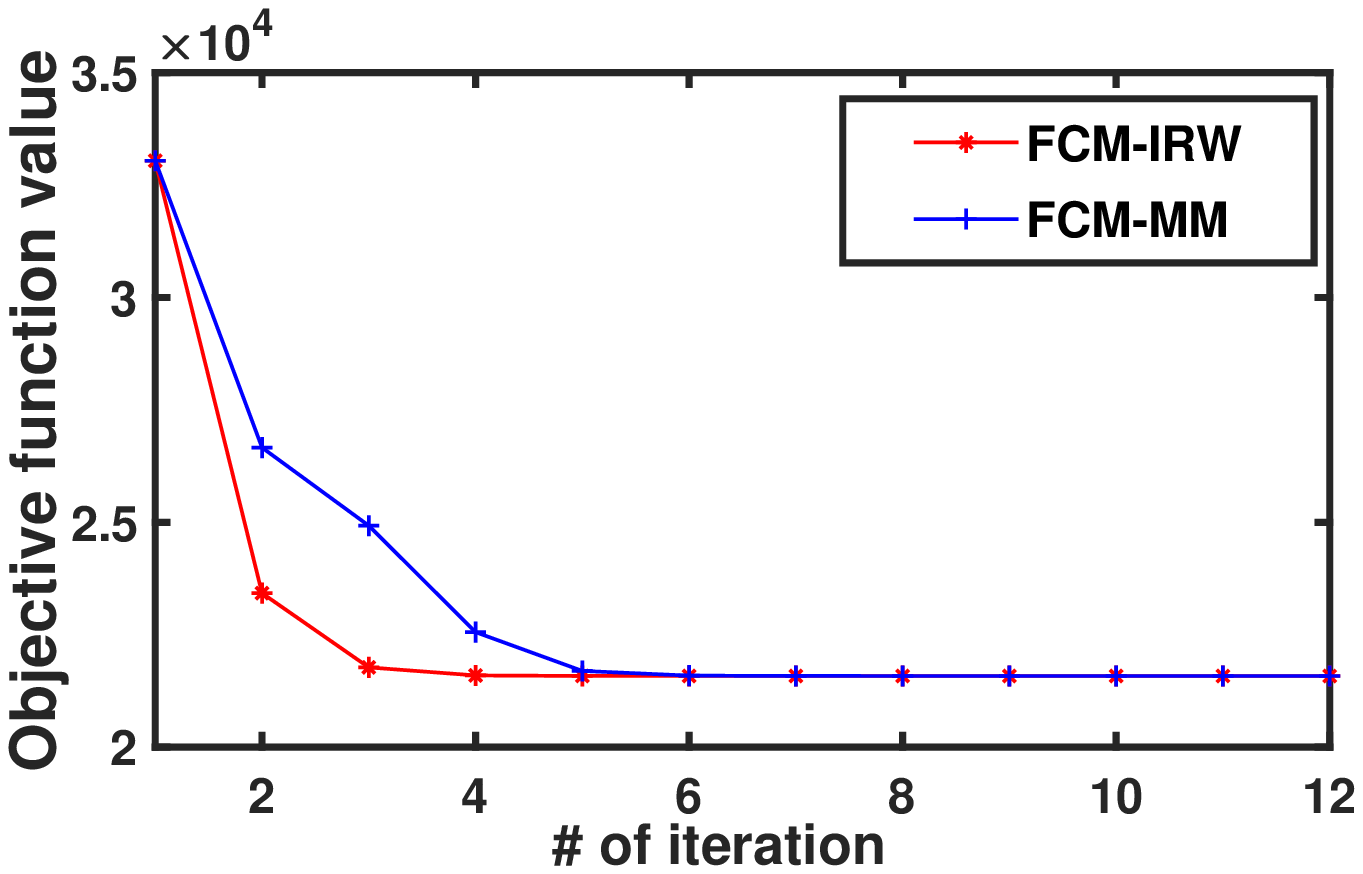}
        \caption{}
        \label{fig:1.d}
    \end{subfigure}
    \hfill
    \begin{subfigure}{0.19\textwidth}
        \centering
        \includegraphics[width=\textwidth,height=1.5cm]{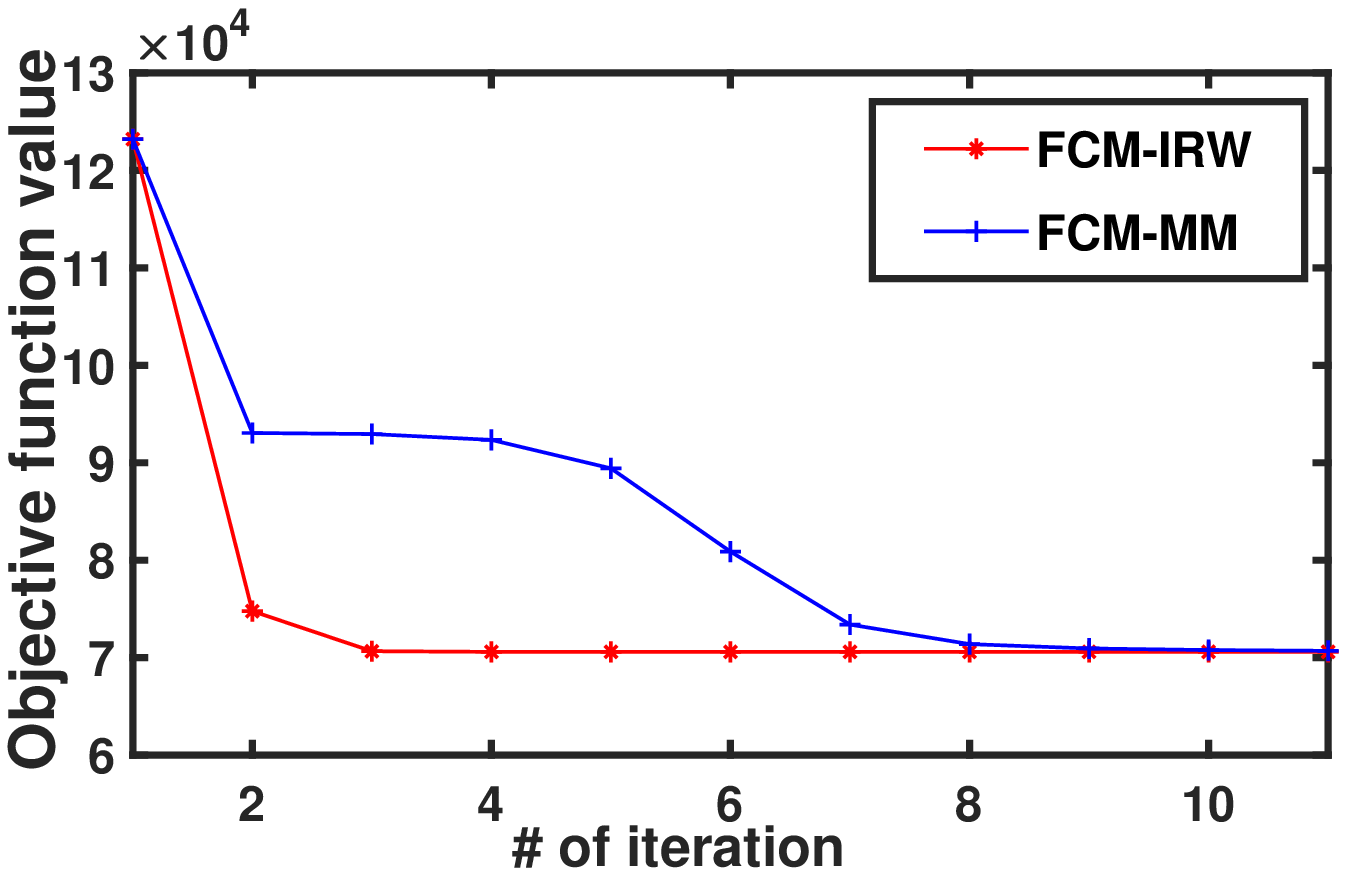}
        \caption{}
        \label{fig:1.e}
    \end{subfigure}
    \newline
    \begin{subfigure}{0.19\textwidth}
        \centering
        \includegraphics[width=\textwidth,height=1.5cm]{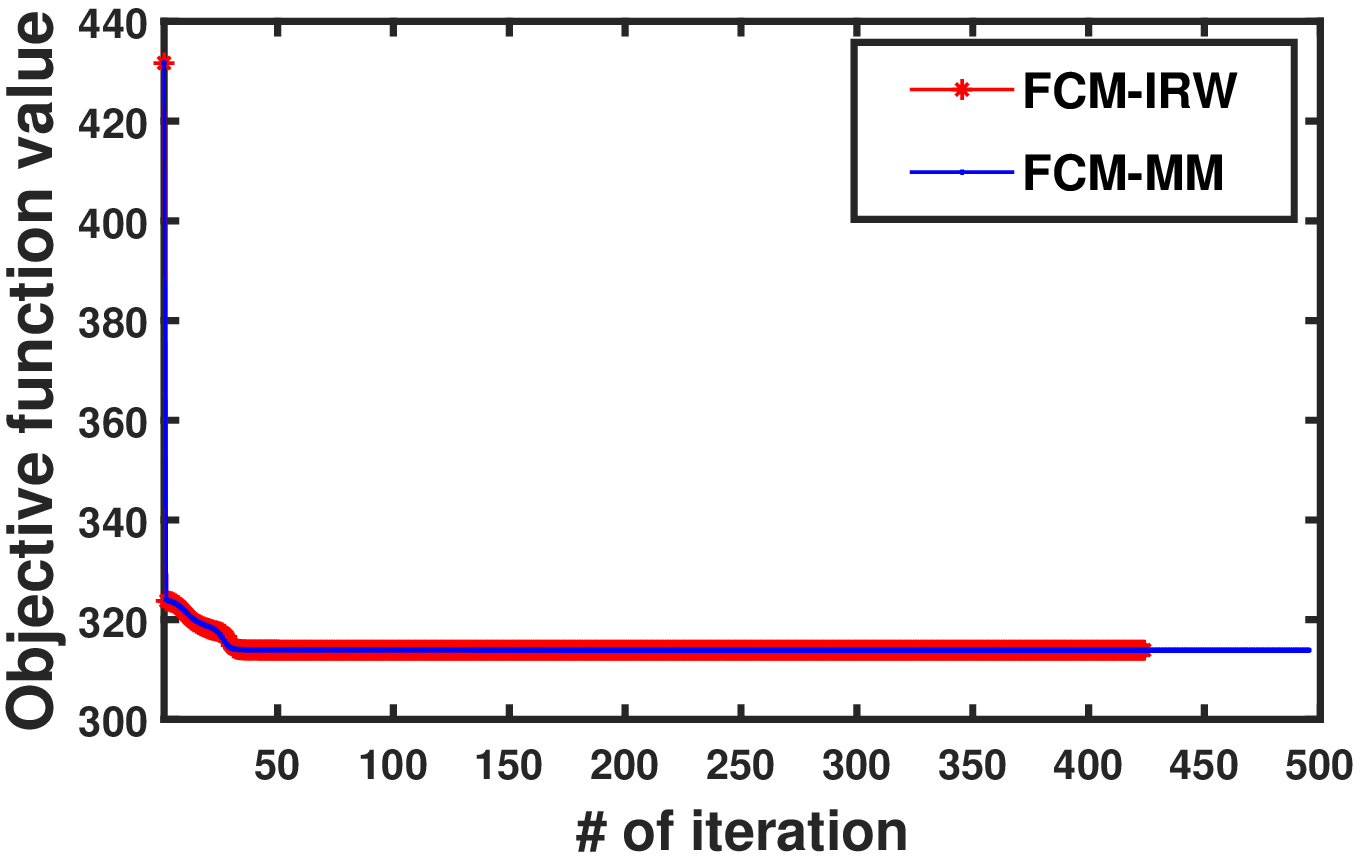}
        \caption{}
        \label{fig:1.f}
    \end{subfigure}
    \hfill
    \begin{subfigure}{0.19\textwidth}
        \centering
        \includegraphics[width=\textwidth,height=1.5cm]{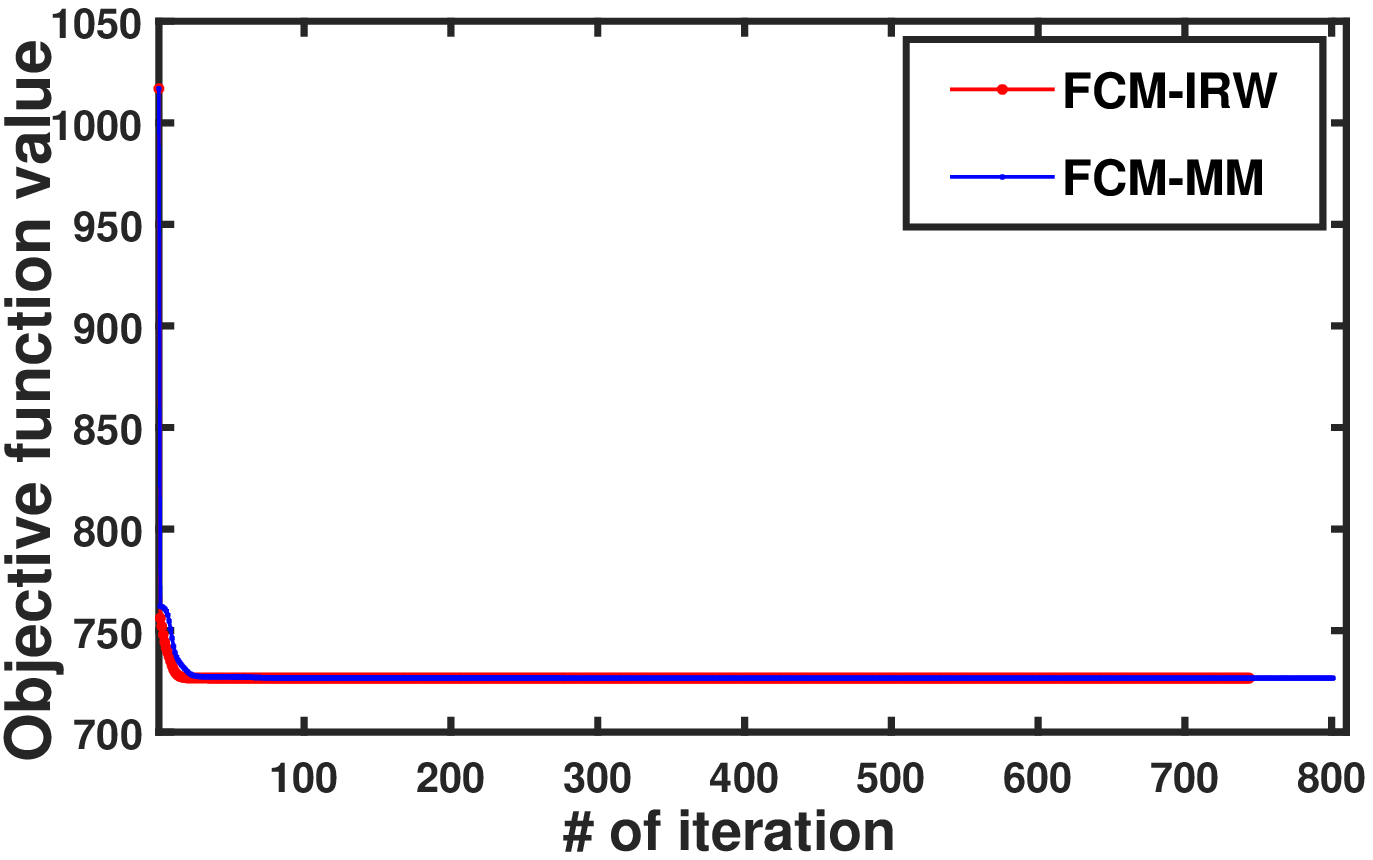}
        \caption{}
        \label{fig:1.g}
    \end{subfigure}
    \hfill
    \begin{subfigure}{0.19\textwidth}
        \centering
        \includegraphics[width=\textwidth,height=1.5cm]{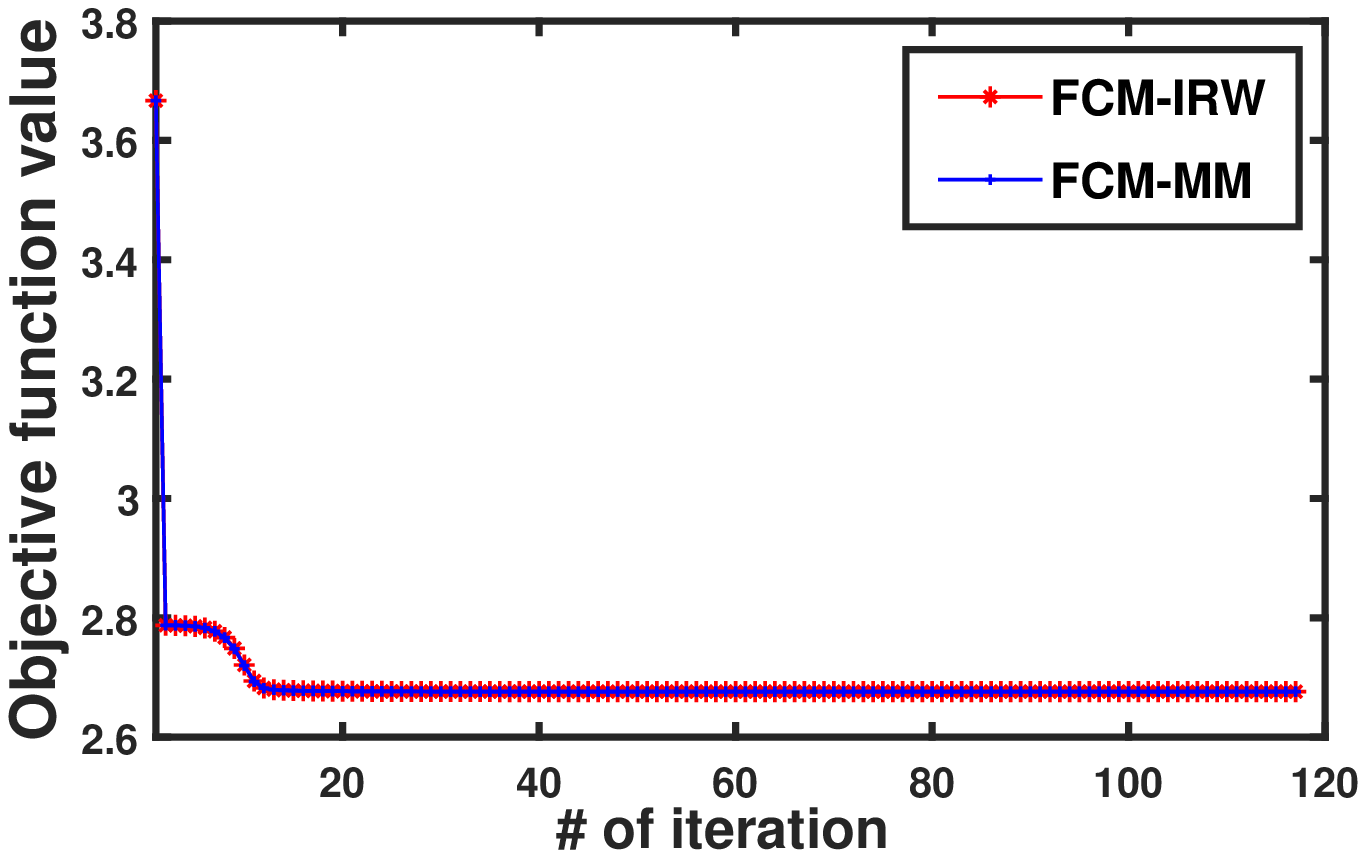}
        \caption{}
        \label{fig:1.h}
    \end{subfigure}
    \hfill
    \begin{subfigure}{0.19\textwidth}
        \centering
        \includegraphics[width=\textwidth,height=1.5cm]{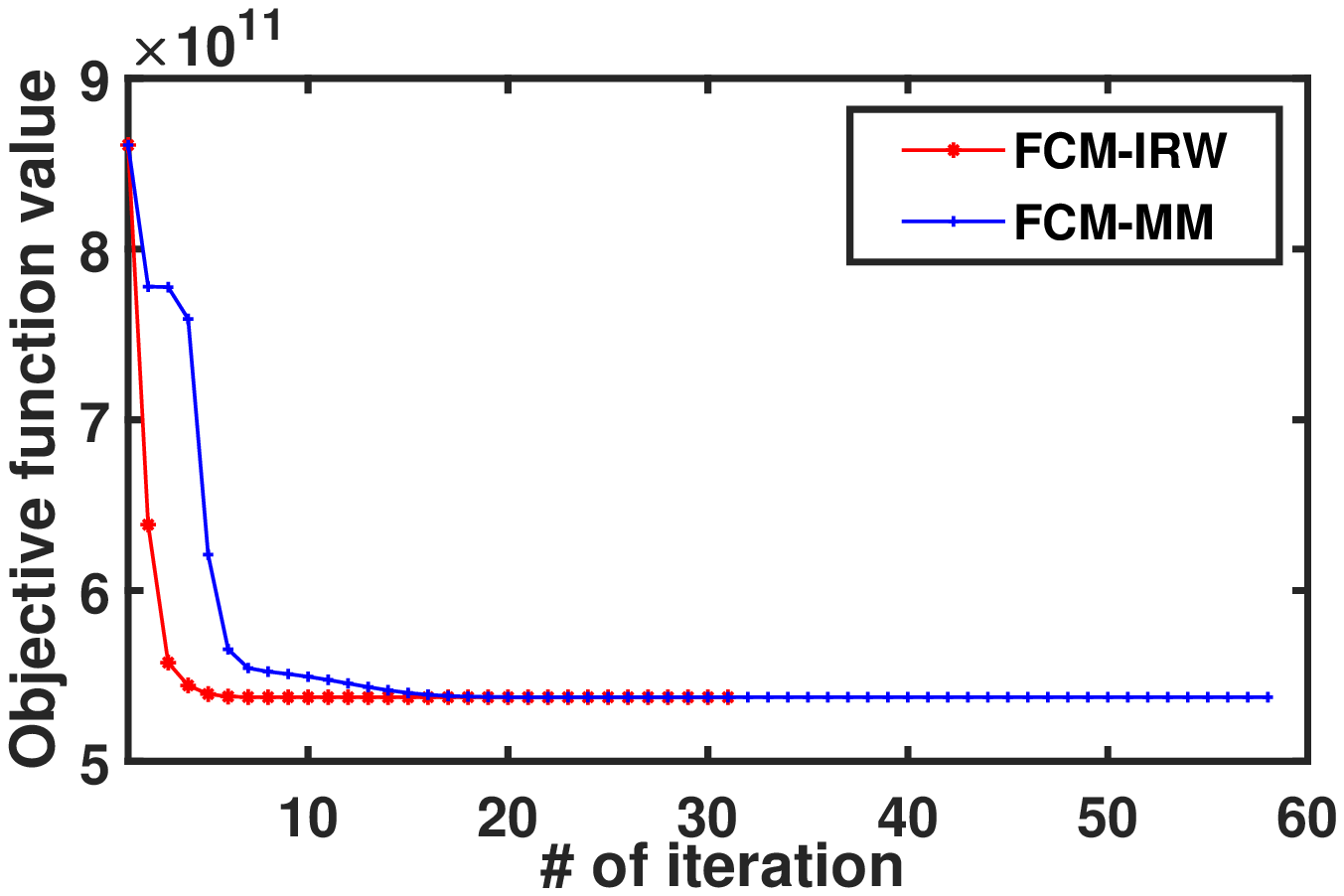}
        \caption{}
        \label{fig:1.i}
    \end{subfigure}
    \hfill
    \begin{subfigure}{0.19\textwidth}
        \centering
        \includegraphics[width=\textwidth,height=1.5cm]{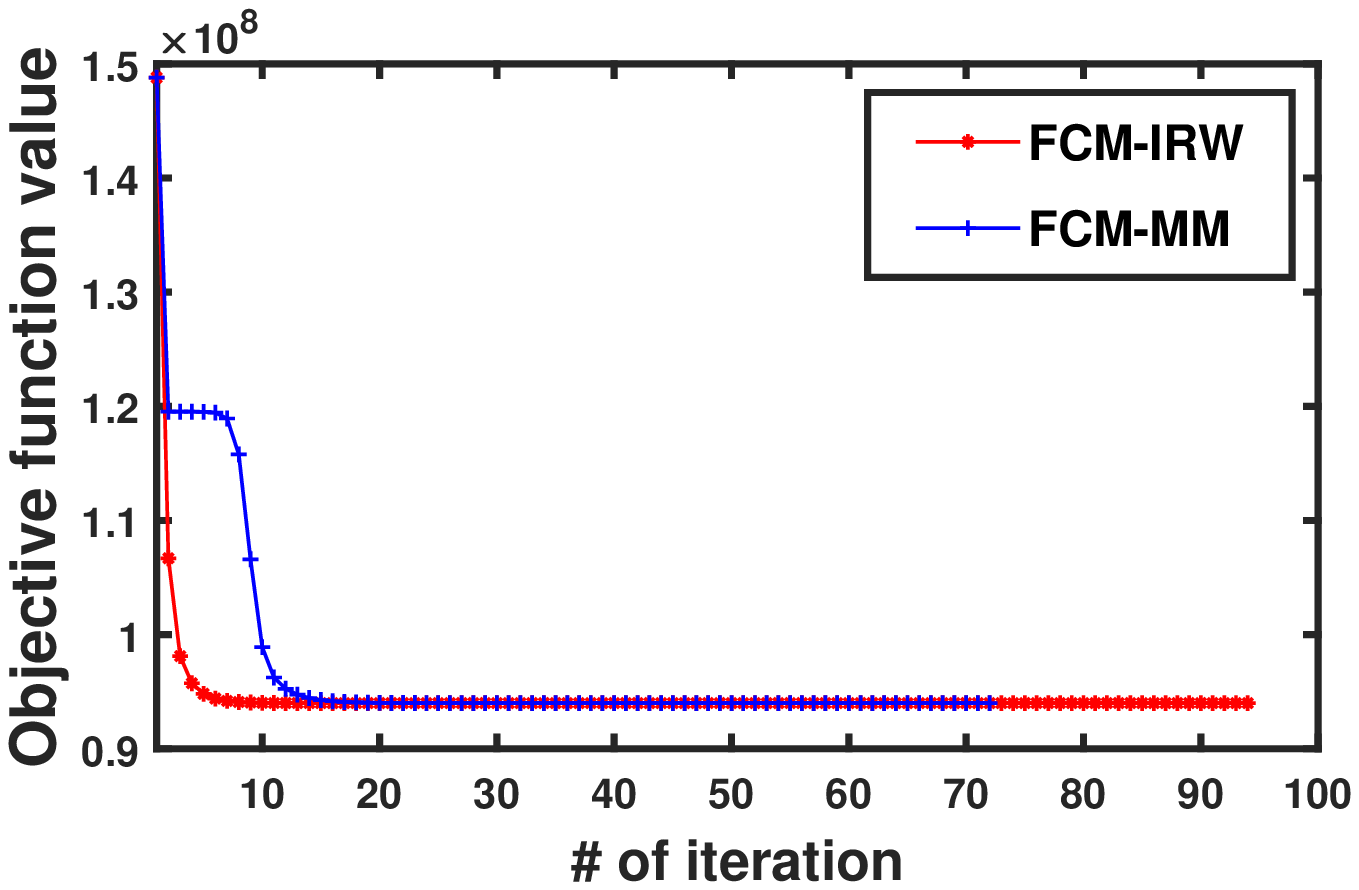}
        \caption{}
        \label{fig:1.j}
    \end{subfigure}
    \hfill
    \caption{Convergence curves for IRW-FCM and FCM-MM with same initialization on 5 datasets (\ref{fig:1.a}) Iris, (\ref{fig:1.b}) Audit, (\ref{fig:1.c}) Automobile, (\ref{fig:1.d}) Haberman Survival, (\ref{fig:1.e}) UCIHAR, (\ref{fig:1.f}) SCADI, (\ref{fig:1.g}) Data Cortex Nuclear, (\ref{fig:1.h}) Gesture Phase  Segmentation (Processed), (\ref{fig:1.i}) EEG Eye State and (\ref{fig:1.j}) HTRU2.}
    \label{fig:Fig1}
\end{figure*}
\begin{figure*}[!h]
\captionsetup{font=small}
\centering
    \begin{subfigure}{0.19\textwidth}
        \centering
        \includegraphics[width=\textwidth,height=1.5cm]{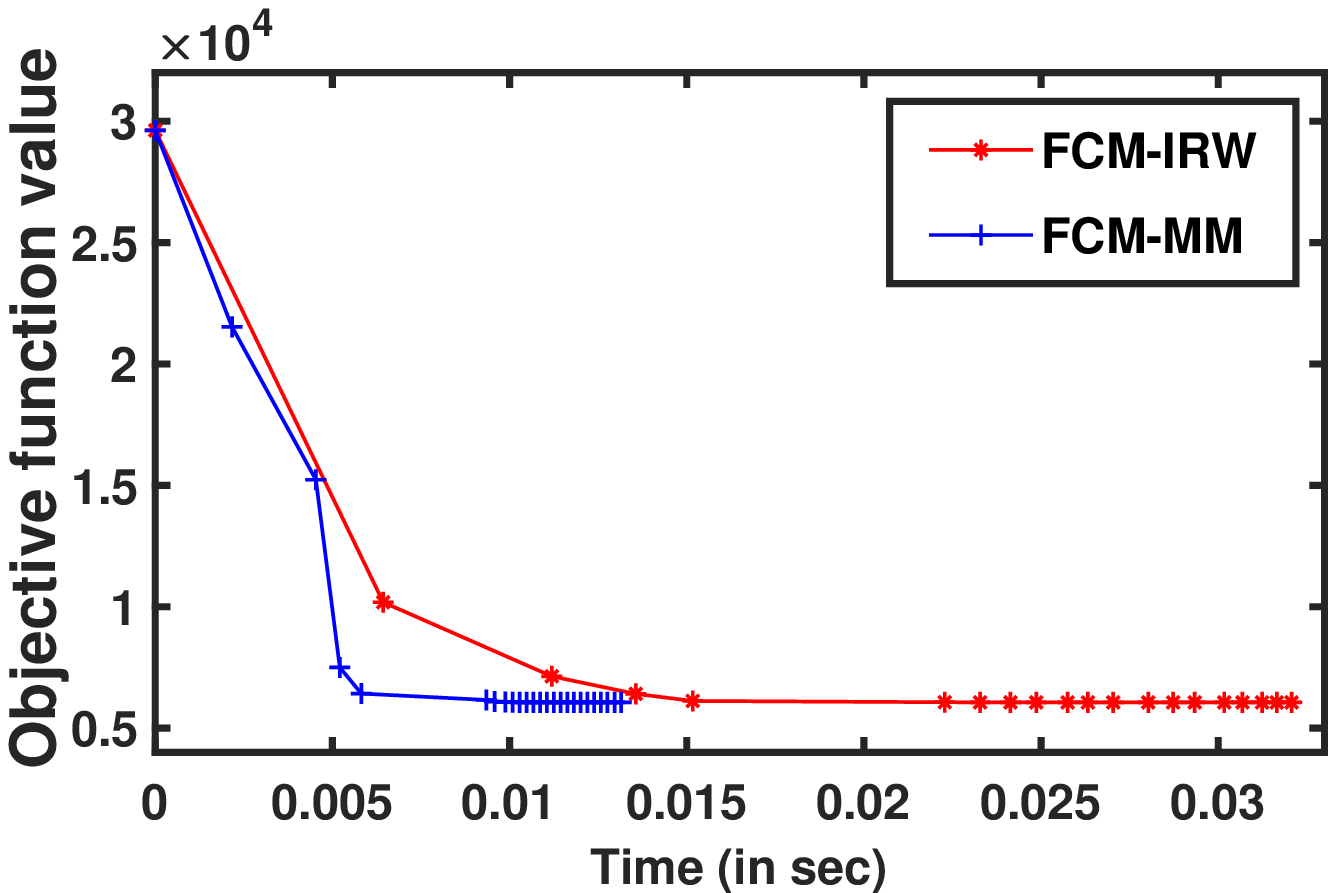}
        \caption{}
        \label{fig:2.a}
    \end{subfigure}
    \hfill
    \begin{subfigure}{0.19\textwidth}
        \centering
        \includegraphics[width=\textwidth,height=1.5cm]{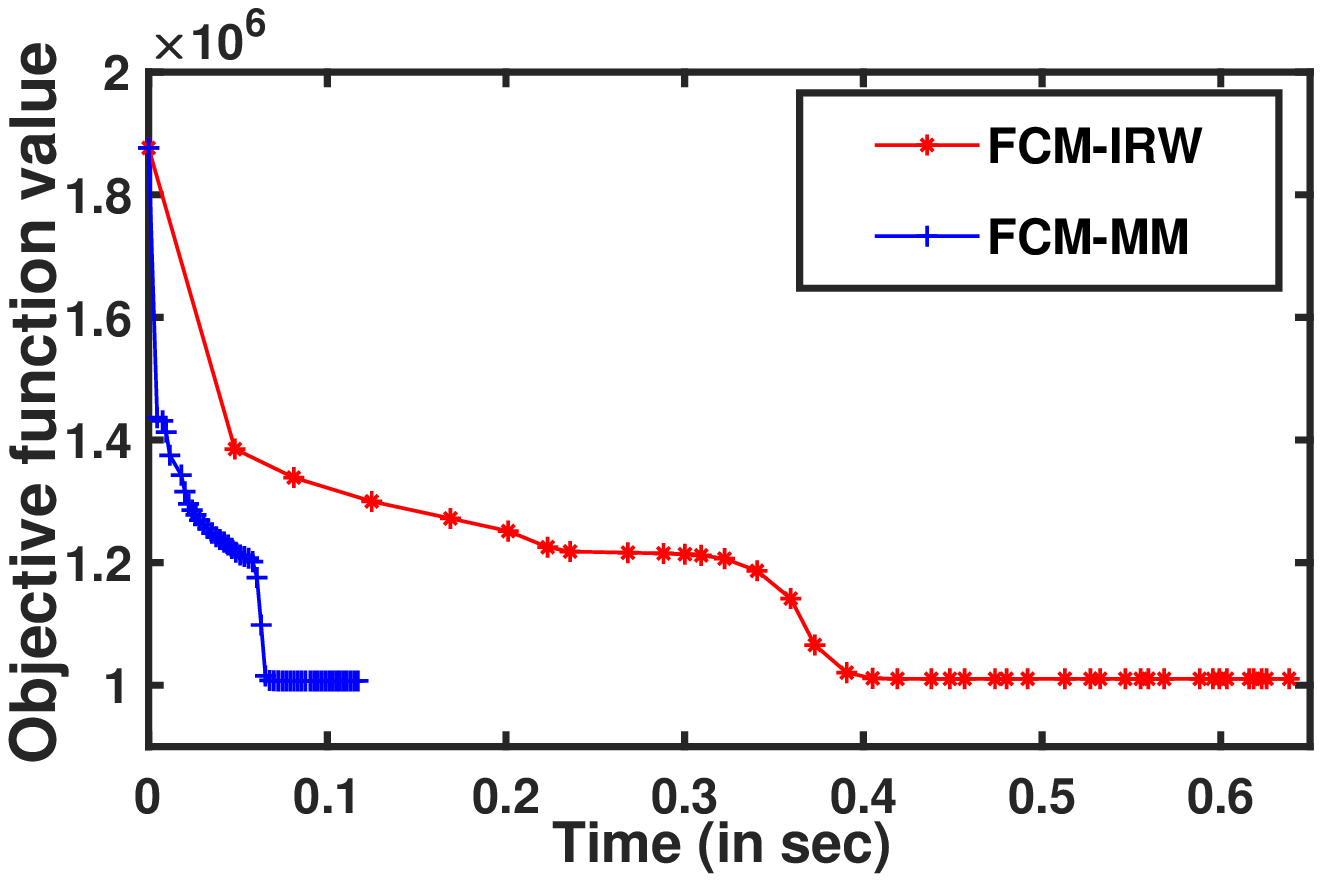}
        \caption{}
        \label{fig:2.b}
    \end{subfigure}
    \hfill
    \begin{subfigure}{0.19\textwidth}
        \centering
        \includegraphics[width=\textwidth,height=1.5cm]{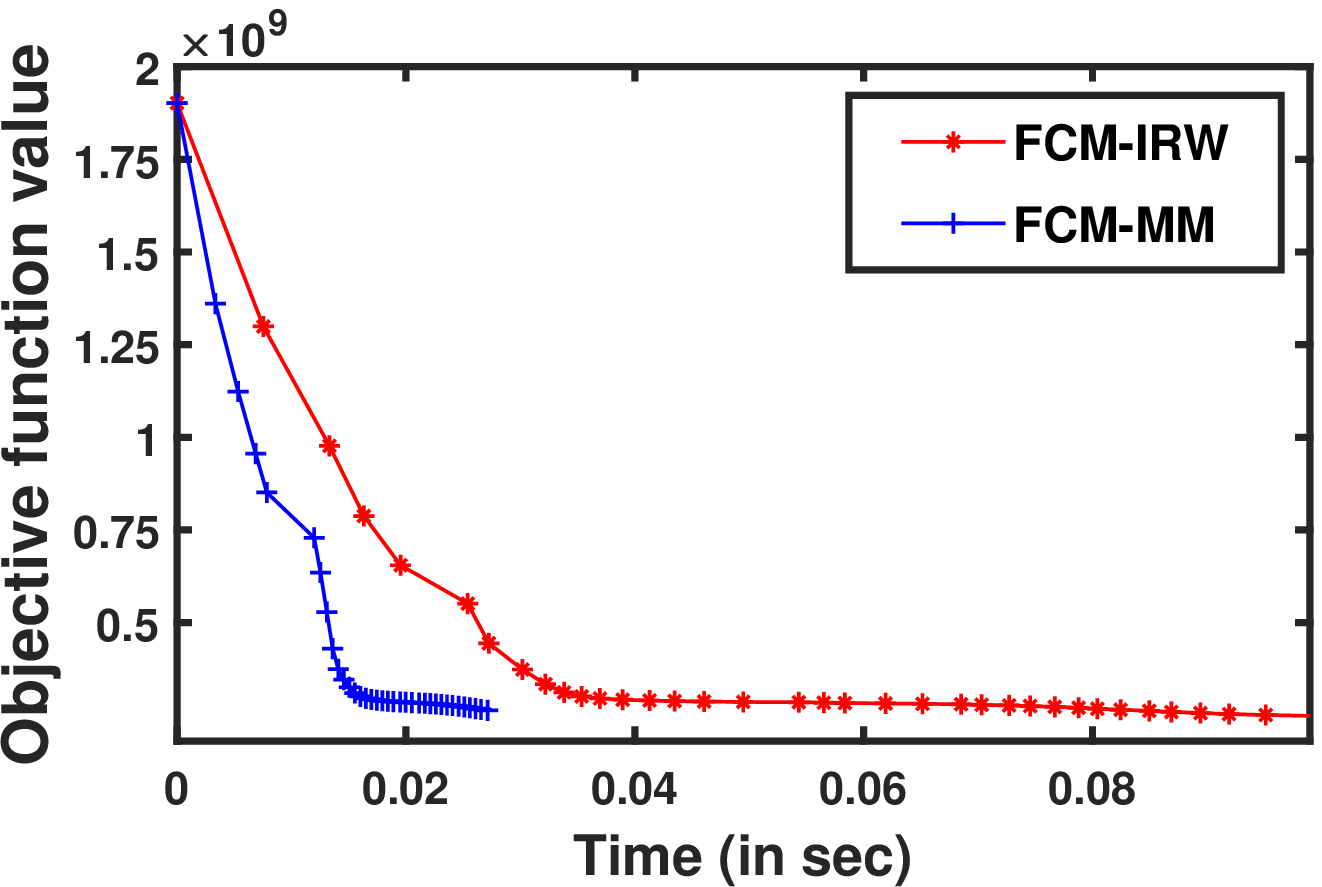}
        \caption{}
        \label{fig:2.c}
    \end{subfigure}
    \hfill
    \begin{subfigure}{0.19\textwidth}
        \centering
        \includegraphics[width=\textwidth,height=1.5cm]{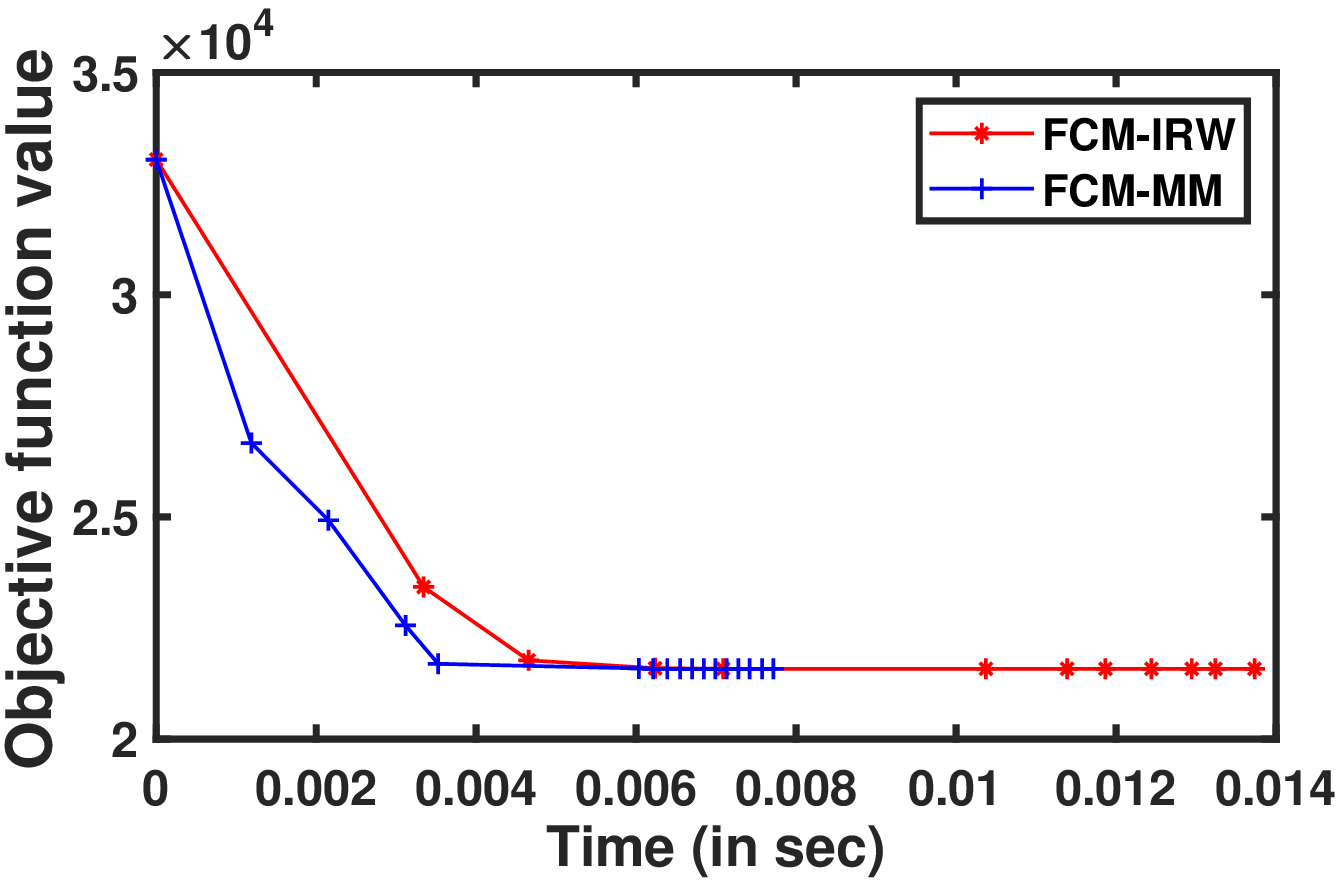}
        \caption{}
        \label{fig:2.d}
    \end{subfigure}
    \hfill
    \begin{subfigure}{0.19\textwidth}
        \centering
        \includegraphics[width=\textwidth,height=1.5cm]{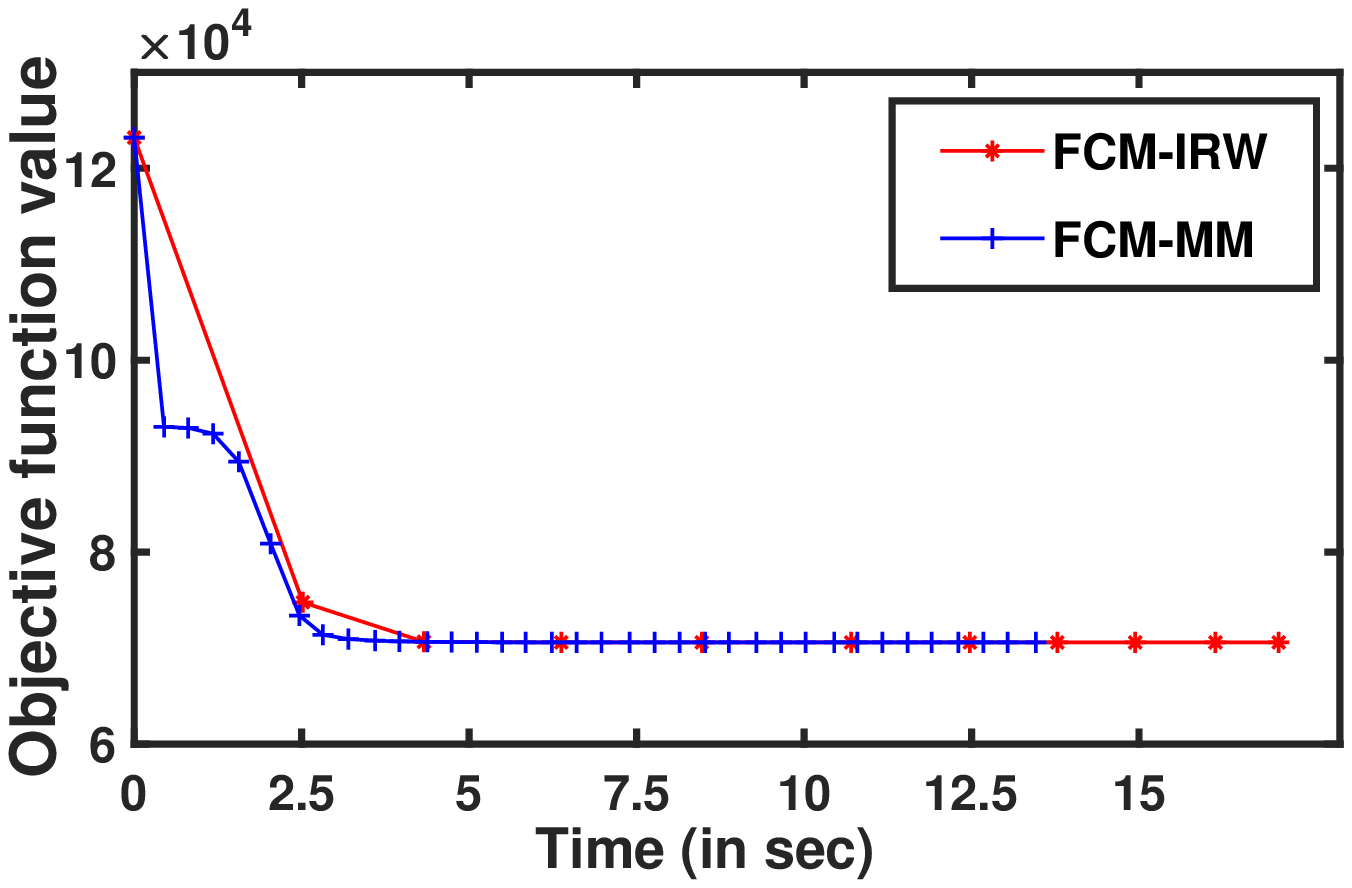}
        \caption{}
        \label{fig:2.e}
    \end{subfigure}
    \newline
    \begin{subfigure}{0.19\textwidth}
        \centering
        \includegraphics[width=\textwidth,height=1.5cm]{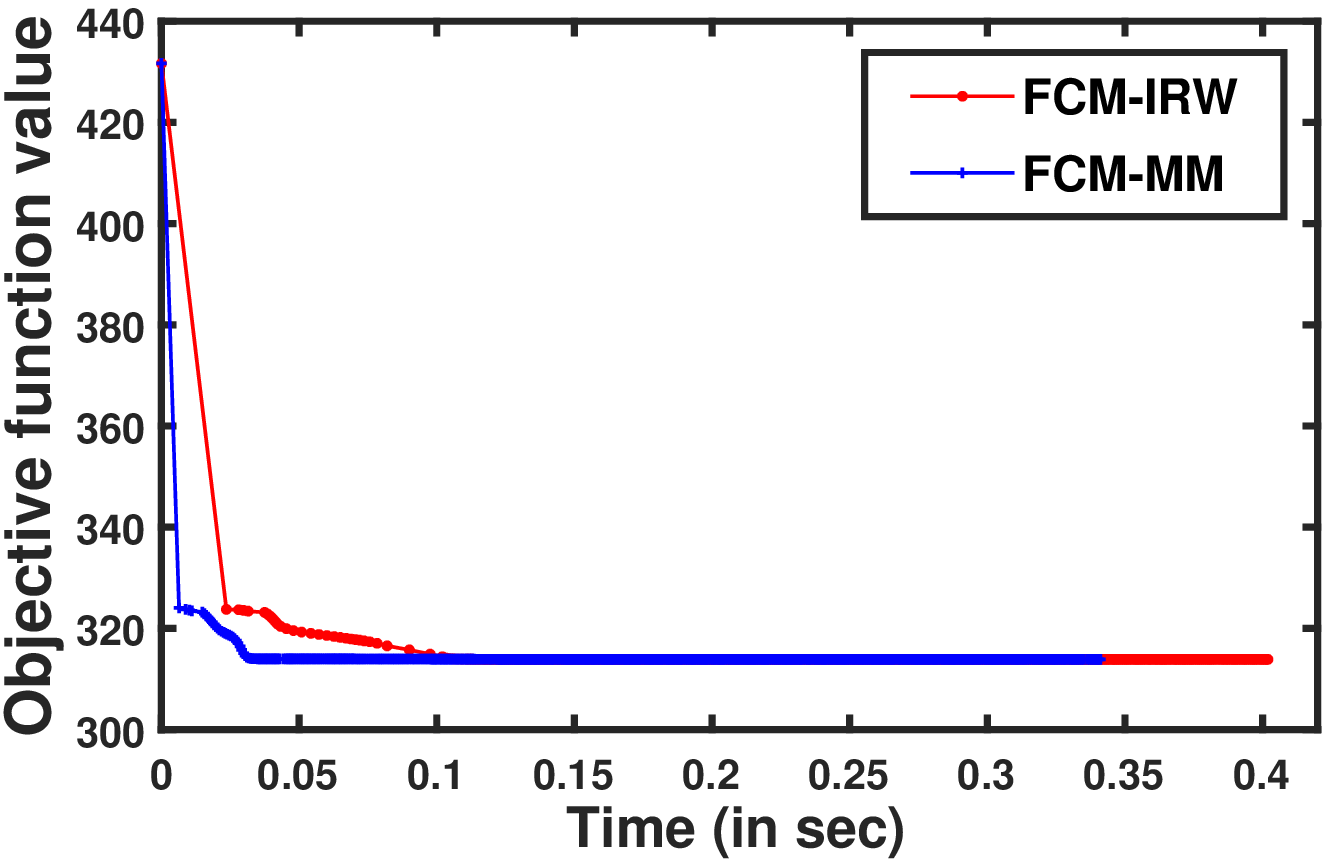}
        \caption{}
        \label{fig:2.f}
    \end{subfigure}
    \hfill
    \begin{subfigure}{0.19\textwidth}
        \centering
        \includegraphics[width=\textwidth,height=1.5cm]{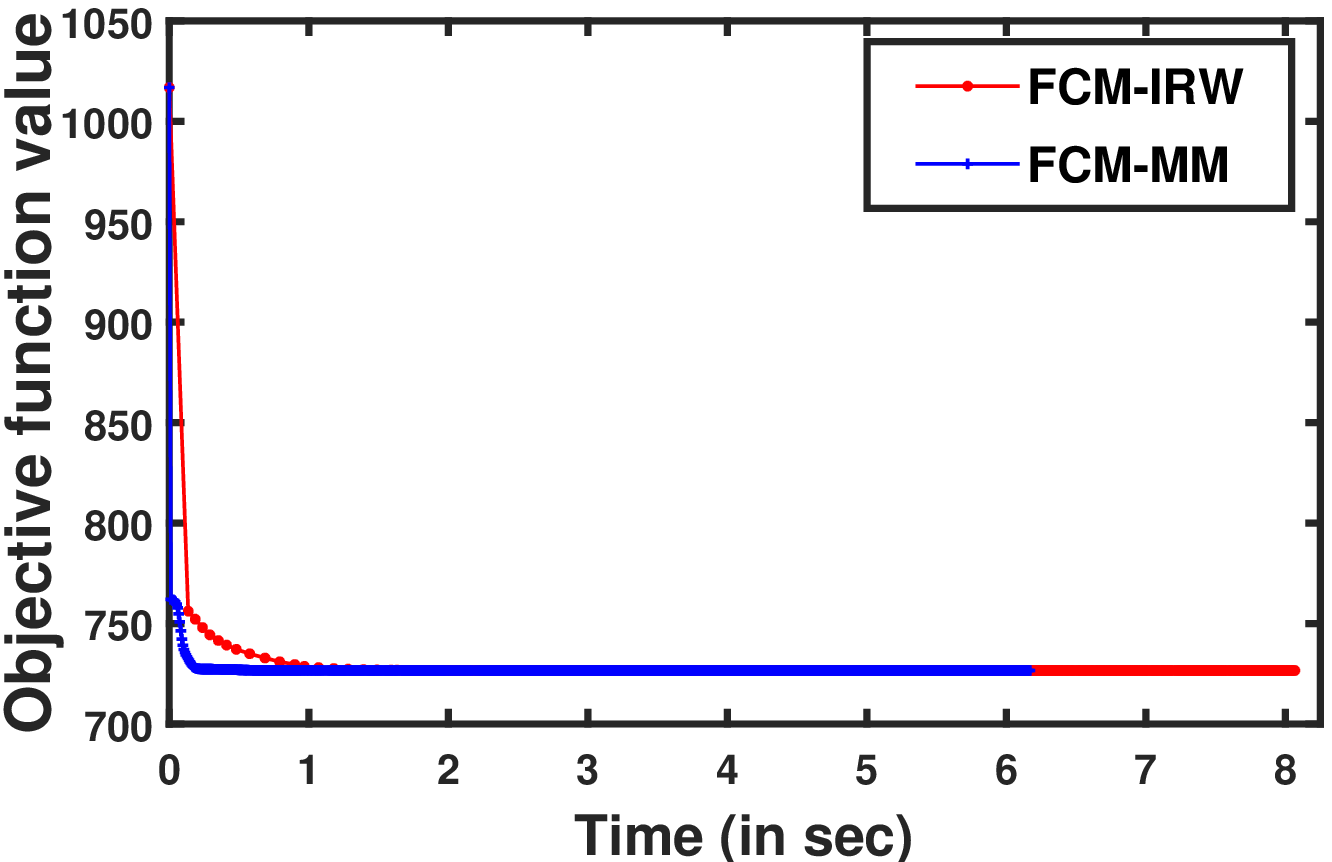}
        \caption{}
        \label{fig:2.g}
    \end{subfigure}
    \hfill
    \begin{subfigure}{0.19\textwidth}
        \centering
        \includegraphics[width=\textwidth,height=1.5cm]{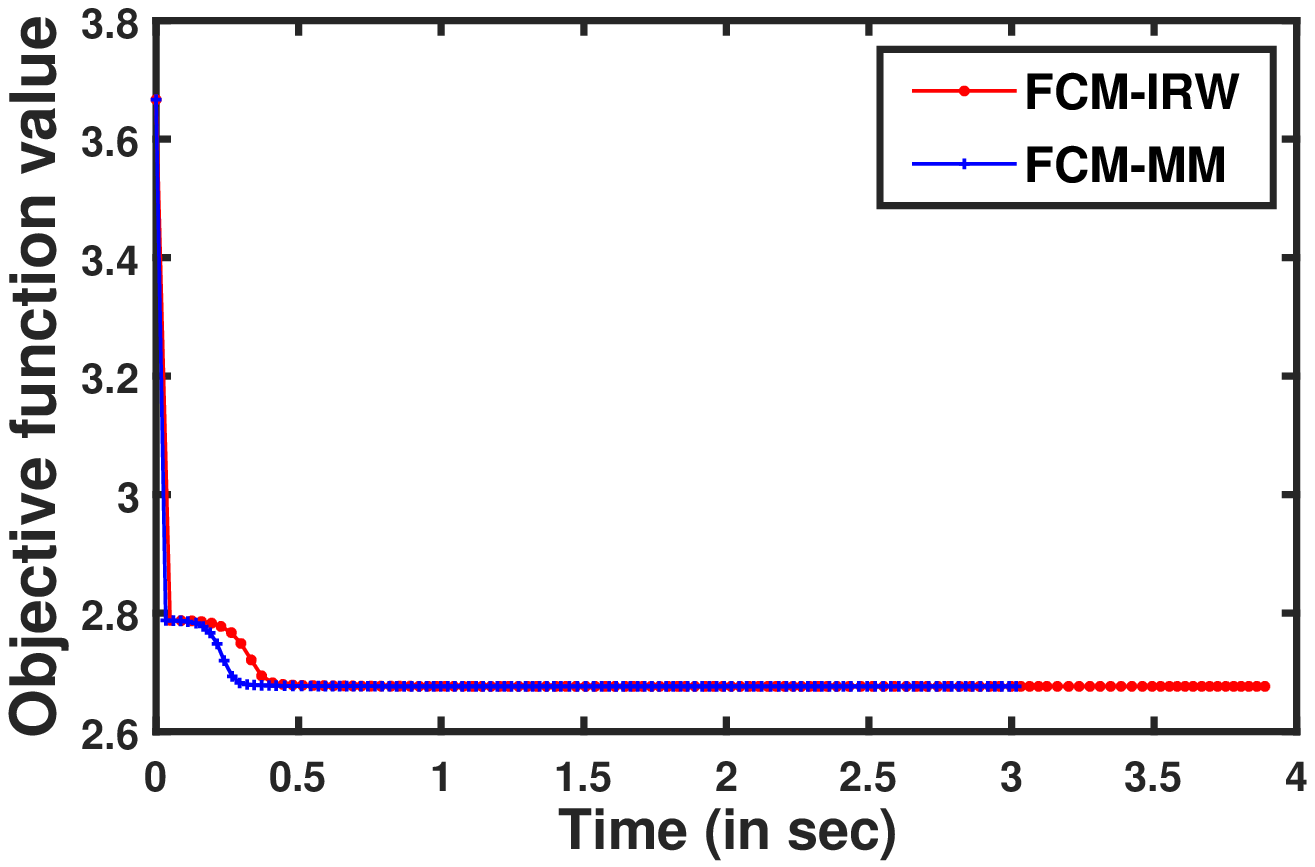}
        \caption{}
        \label{fig:2.h}
    \end{subfigure}
    \hfill
    \begin{subfigure}{0.19\textwidth}
        \centering
        \includegraphics[width=\textwidth,height=1.5cm]{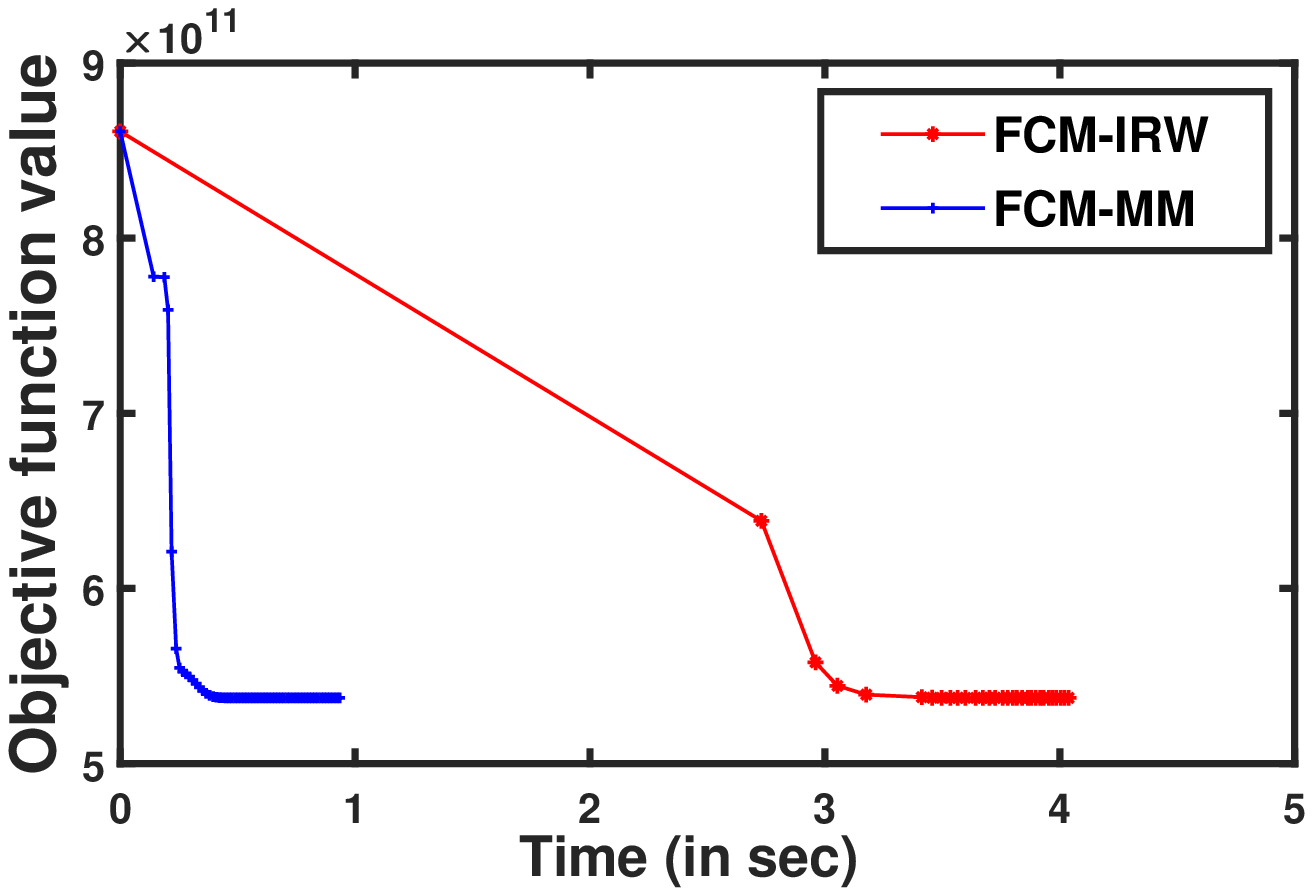}
        \caption{}
        \label{fig:2.i}
    \end{subfigure}
    \hfill
    \begin{subfigure}{0.19\textwidth}
        \centering
        \includegraphics[width=\textwidth,height=1.5cm]{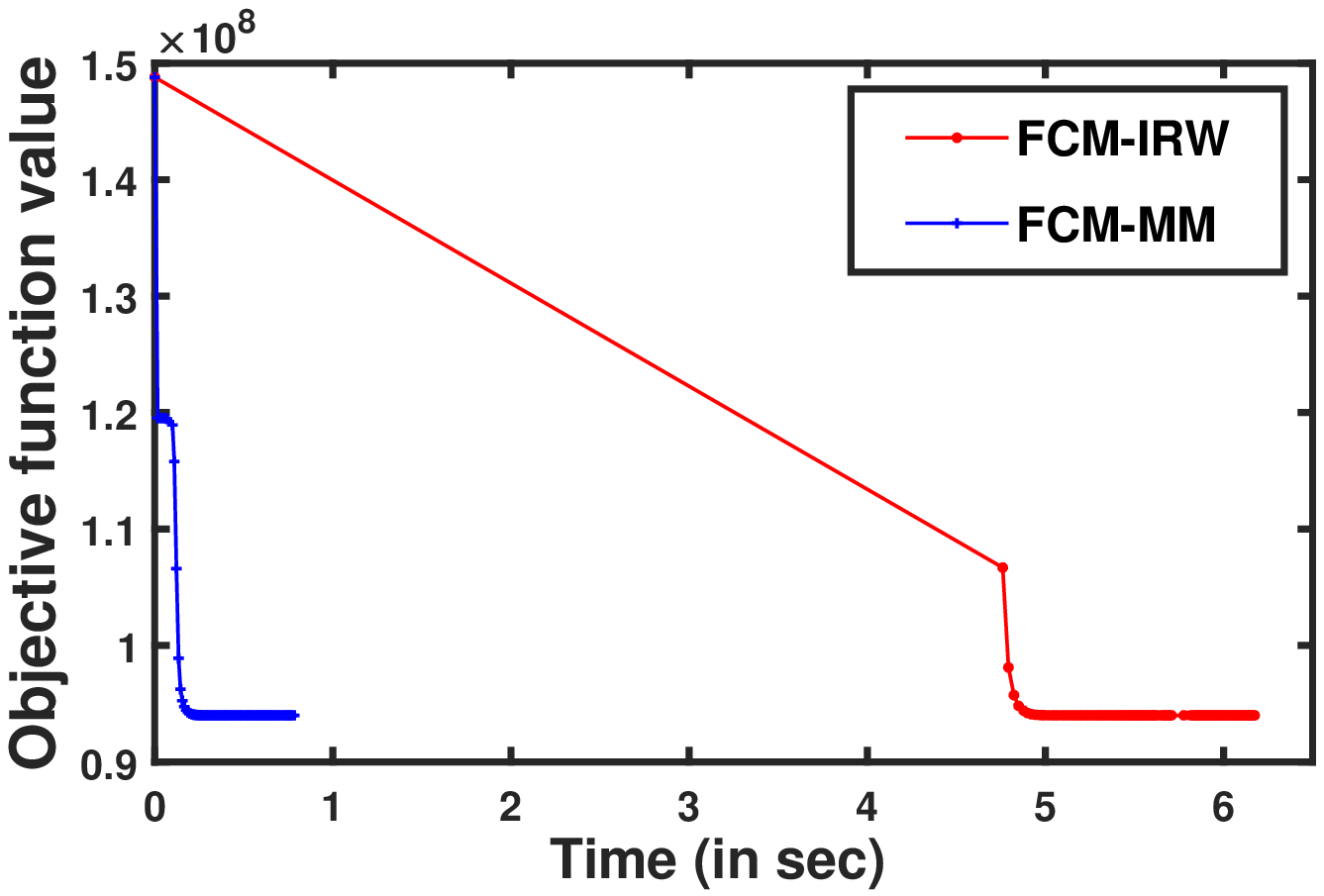}
        \caption{}
        \label{fig:2.j}
    \end{subfigure}
    \caption{Objective function value vs Runtime (Bottom Plot) for IRW-FCM and FCM-MM with same initialization on 5 datasets  (\ref{fig:2.a}) Iris,  (\ref{fig:2.b}) Audit,  (\ref{fig:2.c}) Automobile, (\ref{fig:2.d}) Haberman Survival,  (\ref{fig:2.e}) UCIHAR, (\ref{fig:2.f}) SCADI, (\ref{fig:2.g}) Data Cortex Nuclear,  (\ref{fig:2.h}) Gesture Phase  Segmentation (Processed),  (\ref{fig:2.i}) EEG Eye State and  (\ref{fig:2.j}) HTRU2.}
    \label{fig:Fig2}
\end{figure*}
In this section, we present some simulation results comparing the double loop IRW-FCM\footnote[1]{The MATLAB codes for IRW-FCM are obtained from https://github.com/Sara-Jingjing-Xue} from \cite{ref1} and the MM based algorithm derived in this note (we named it as FCM-MM). We consider five different datasets from the UCI ML Repository\footnote[2]{http://archive.ics.uci.edu/ml/index.php} with features mentioned in Table \ref{table:tab1}. In Figure \ref{fig:Fig1} and  Figure \ref{fig:Fig2}, we show the objective vs iteration and objective vs runtime (in sec) of the IRW-FCM and FCM-MM algorithms for the five datasets. The initialization ($\mathbf{F}^0$) for both the algorithms is kept the same and in the case of IRW-FCM, the objective is plotted with respect to the outer iterations and the inner loop is run till convergence. It can be noticed from figure \ref{fig:Fig1} that IRW-FCM seem to be converging faster in terms of iterations, however, in terms of runtime FCM-MM is generally faster than IRW-FCM algorithm- this clearly indicates that running the inner loop of IRW-FCM till convergence is not necessary and it can be run only one time to attain the same minimum at a much faster runtime.
\vspace{-0.4cm}
\section{Conclusion}
In this note, we present a simple alternate derivation to the IRW-FCM algorithm presented in \cite{ref1} to solve the Fuzzy c-Means problem. The derivation presented in this note is based on the popular Majorization-Minimization approach which enjoys nicer convergence properties. The derivation presented in this note shows that the inner loop in the IRW-FCM algorithm can be eliminated (by running it only one time) and yet convergence can be guaranteed.
\vspace{-0.4cm}
\bibliographystyle{IEEEtran}
\bibliography{reference_fcm_mm}


\newpage
\vspace{11pt}

\vfill

\end{document}